\documentclass[10pt,twocolumn,letterpaper]{article}

\usepackage{iccv}
\usepackage[T1]{fontenc}
\usepackage{times}
\usepackage{epsfig}
\usepackage{graphicx}
\usepackage{amsmath}
\usepackage{amssymb}
\usepackage{microtype}
\usepackage{booktabs}
\usepackage{multirow}
\usepackage{caption}
\usepackage{subcaption}
\usepackage{bbm}
\usepackage{pifont}
\usepackage{enumitem}
\usepackage[super]{nth}
\usepackage[dvipsnames]{xcolor, mathtools}
\usepackage[ruled,vlined, noend]{algorithm2e}
\usepackage[accsupp]{axessibility}

\definecolor{citecolor}{HTML}{0071bc}
\definecolor{bluegreen}{RGB}{117, 171, 188}
\definecolor{mybrown}{HTML}{D7C0AE}

\usepackage[pagebackref,breaklinks,colorlinks, anchorcolor=blue, citecolor=citecolor]{hyperref}

\usepackage[capitalize]{cleveref}
\crefname{section}{Sec.}{Secs.}
\Crefname{section}{Section}{Sections}
\Crefname{table}{Table}{Tables}
\crefname{table}{Tab.}{Tabs.}

\newcommand{\cmark}{\ding{51}}%
\newlength\savewidth
\newcommand{\colorhat}[1]{{\color{red}\hat{\color{black}{#1}}}}

\iccvfinalcopy 

\def\iccvPaperID{9001} 

\ificcvfinal\fi

\begin{document}

\title{Shrinking Class Space for Enhanced Certainty in Semi-Supervised Learning}

\author{Lihe Yang$^1$~~~~Zhen Zhao$^4$~~~~Lei Qi$^5$~~~~Yu Qiao$^3$~~~~Yinghuan Shi$^2$\hspace{0.2mm}$^*$~~~~Hengshuang Zhao$^1$\thanks{Corresponding authors}\vspace{1mm}\\
$^1$The University of Hong Kong~~~~$^2$Nanjing University\\$^3$Shanghai AI Laboratory~~~~$^4$The University of Sydney~~~~$^5$Southeast University\\
\small{\url{https://github.com/LiheYoung/ShrinkMatch}}
}

\maketitle
\ificcvfinal\thispagestyle{empty}\fi

\begin{abstract}
    Semi-supervised learning is attracting blooming attention, due to its success in combining unlabeled data. To mitigate potentially incorrect pseudo labels, recent frameworks mostly set a fixed confidence threshold to discard uncertain samples. This practice ensures high-quality pseudo labels, but incurs a relatively low utilization of the whole unlabeled set. In this work, our key insight is that these uncertain samples can be turned into certain ones, as long as the confusion classes for the top-1 class are detected and removed. Invoked by this, we propose a novel method dubbed ShrinkMatch to learn uncertain samples. For each uncertain sample, it adaptively seeks a shrunk class space, which merely contains the original top-1 class, as well as remaining less likely classes. Since the confusion ones are removed in this space, the re-calculated top-1 confidence can satisfy the pre-defined threshold. We then impose a consistency regularization between a pair of strongly and weakly augmented samples in the shrunk space to strive for discriminative representations. Furthermore, considering the varied reliability among uncertain samples and the gradually improved model during training, we correspondingly design two reweighting principles for our uncertain loss. Our method exhibits impressive performance on widely adopted benchmarks.
\end{abstract}

\section{Introduction}

In the last decade, our computer vision community has witnessed inspiring progress, thanks to large-scale datasets \cite{coco, citys}. Nevertheless, it is laborious and costly to annotate massive images, hindering the progress to benefit a broader range of real-world scenarios. Inspired by this, semi-supervised learning (SSL) was proposed to utilize the unlabeled data under the assistance of limited labeled data.

\begin{figure}
    \centering
    \includegraphics[width=0.8\linewidth]{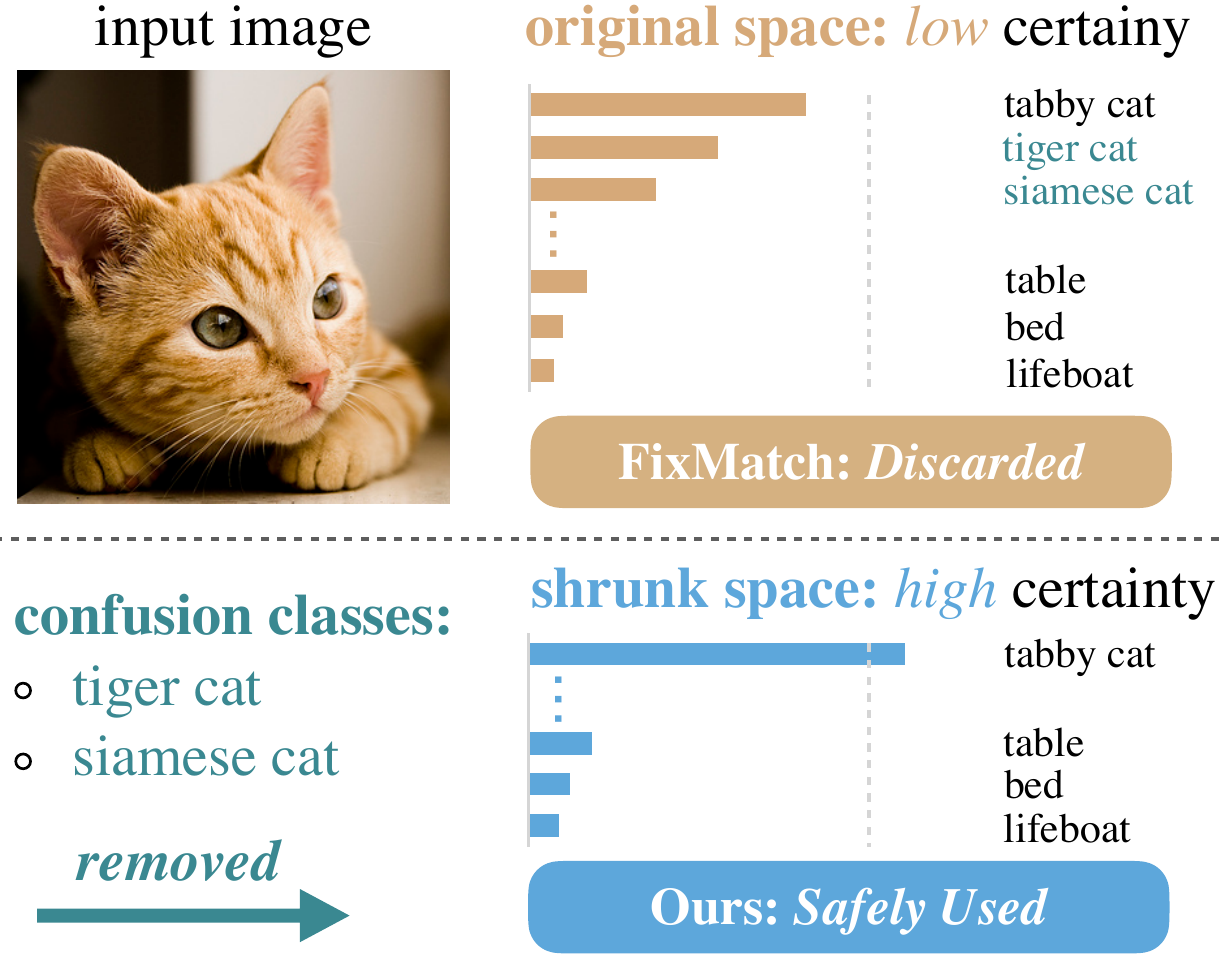}
    \caption{Illustration of our motivation. Due to confusion classes for the top-1 class, the certainty fails to reach the pre-defined threshold \textcolor{gray}{(gray dotted line)}. FixMatch discards such uncertain samples. Our method, however, detects and removes confusion classes to enhance certainty, then enjoying full and safe utilization of all unlabeled images.}
    \label{fig:intro}
\vspace{-4mm}
\end{figure}

The frameworks in SSL are typically based on the strategy of pseudo labeling. Briefly, the model acquires knowledge from the labeled data, and then assigns predictions on the unlabeled data. The two sources of data are finally combined to train a better model. During this process, it is obvious that predictions on unlabeled data are not reliable. If the model is iteratively trained with incorrect pseudo labels, it will suffer the confirmation bias issue \cite{arazo2020pseudo}. To address this dilemma, recent works \cite{fixmatch} simply set a fixed confidence threshold to discard potentially unreliable samples. This simple strategy effectively retains high-quality pseudo labels, however, it also incurs a low utilization of the whole unlabeled set. As evidenced by our pilot study on CIFAR-100 \cite{cifar}, nearly 20\% unlabeled images are filtered out for not satisfying the threshold of 0.95. Instead of blindly throwing them away, we believe there should exist a more promising approach. This work is just aimed to fully leverage previously uncertain samples in an \emph{informative} but also \emph{safe} manner.

So first, why are these samples uncertain? According to our observations on CIFAR-100 and ImageNet, although the top-1 accuracy could be low, the top-5 accuracy is much higher. This indicates in most cases, the model struggles to discriminate among a small portion of classes. As illustrated in \cref{fig:intro}, given a cat image, the model is not sure whether it belongs to \texttt{tabby} \texttt{cat}, \texttt{tiger} \texttt{cat}, or other cats. On the other hand, however, it is absolutely certain that the object is more like a \texttt{tabby} \texttt{cat}, rather than a \texttt{table} or anything else. In other words, it is reliable for the model to distinguish the top class from the remaining less likely classes.

\begin{figure}
    \centering
    \includegraphics[width=0.85\linewidth]{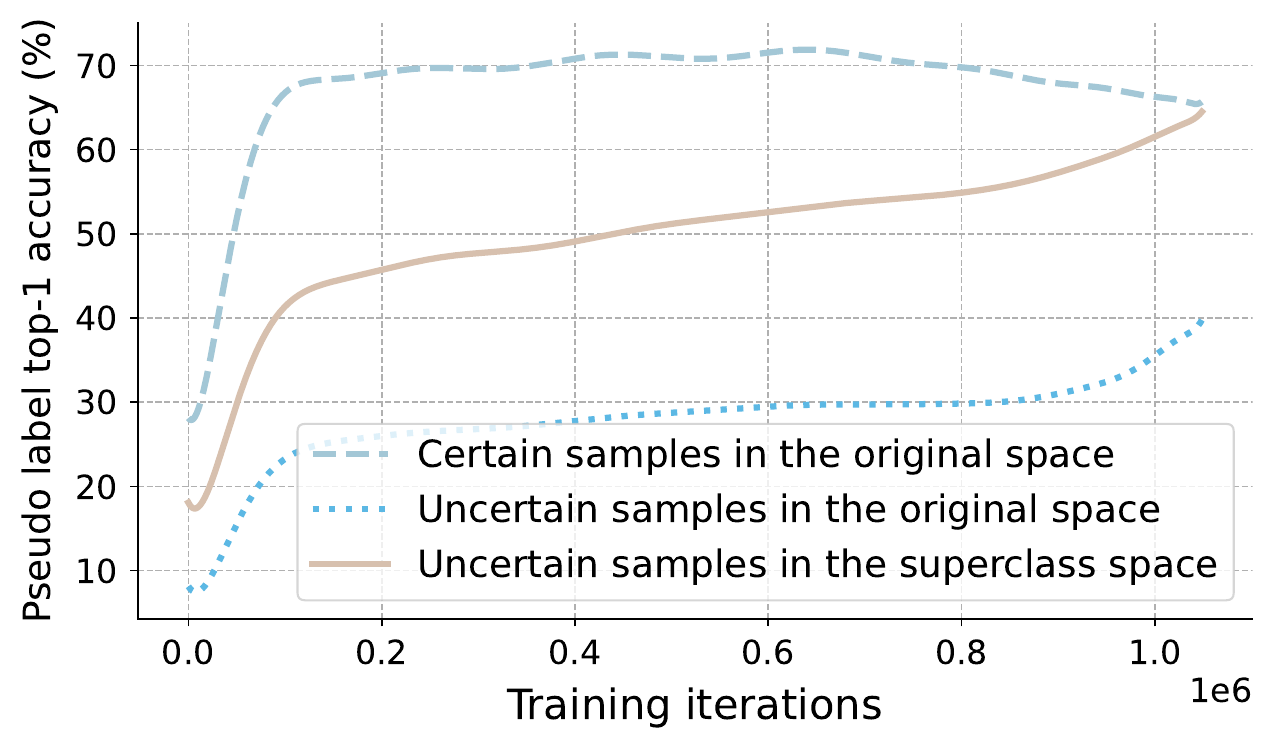}
    \caption{Pseudo label accuracy on CIFAR-100 with 400 labels. We highlight that even for uncertain samples, their top-1 predictions are of high accuracy \emph{in the superclass space} (20 classes). This accuracy can even be comparable to the delicately selected certain samples in the original class space.}
    \vspace{-2mm}
    \label{fig:superclass}
\vspace{-4mm}
\end{figure}

Invoked by these, we propose a novel method dubbed ShrinkMatch to learn uncertain samples. The prediction fails to satisfy the pre-defined threshold due to the existence of confusion classes for the top-1 class. Hence, our approach adaptively seeks a shrunk class space where the confusion classes are removed, to enable the re-calculated confidence to reach the original threshold. Moreover, the obtained shrunk class space is also required to be the largest among those that can satisfy the threshold. In a word, we seek a \emph{certain and largest shrunk space} for uncertain samples. Then, logits of the strongly augmented image are correspondingly gathered in the new space. And a consistency regularization is imposed between the shrunk weak-strong predictions. 

Note that, the confusion classes are detected and removed in a fully automatic and instance-adaptive fashion. Moreover, even if the predicted top-1 class is not fully in line with the groundtruth label, they mostly belong to the same superclass. To prove this, as shown in \cref{fig:superclass}, uncertain samples exhibit much higher pseudo label accuracy in the superclass space than the original space. This accuracy is even comparable to that of certain samples in the original class space.  Therefore, contrasting these groundtruth-related classes against remaining unlikely classes is still highly beneficial, yielding more discriminative representations of our model. To avoid affecting the main classifier, we further adopt an auxiliary classifier to disentangle the learning in the shrunk space.

Despite the effectiveness, there still exist two main drawbacks to the above optimization target. (1) First, it treats all uncertain samples equally. The truth, however, is that in the original class space, the top-1 confidence of different uncertain samples can vary dramatically. And it is clear that samples with larger confidence should be attached more importance. To this end, we propose to balance different uncertain samples by their confidence in the original space. (2) Moreover, the regularization term also overlooks the gradually improved model state during training. At the start of training, there are abundant uncertain samples, but their predictions are extremely noisy, or even random. So even the highest scored class may share no relationship with the true class. Then as the training proceeds, the top classes become reliable. Considering this, we further propose to adaptively reweight the uncertain loss according to the model state. The model state is tracked and approximately estimated via performing exponential moving average on the proportion of certain samples in each mini-batch. With the two reweighting principles, the model turns out more stable, and avoids accumulating much noise from uncertain samples, especially at early training iterations.

To summarize, our contributions lie in three aspects:

\begin{itemize}
    \item We first point out that low certainty is typically caused by a small portion of confusion classes. To enhance the certainty, we propose to shrink the original class space by adaptively detecting and removing confusion ones for the top-1 class to turn it certain in the new space.
    \item We manage to reweight the uncertain loss from two perspectives: the image-based varied reliability among different uncertain samples, and the model-based gradually improved state as the training proceeds.
    \item Our proposed ShrinkMatch establishes new state-of-the-art results on widely acknowledged benchmarks.
\end{itemize}
\section{Related Work}

\noindent
\textbf{Semi-supervised learning (SSL).} The primary concern in SSL \cite{pseudolabel, fixmatch, dash, comatch, meta, conmatch, yang2022class, noisystudent, s4l, lassl, simple, overviewssl, taherkhani2021self, nassar2021all, cai2022semi, tang2022towards, fan2022cossl, wang2022unsupervised, assran2021semi, flexmatch} is to design effective constraints for unlabeled samples. Dating back to decades ago, pioneering works \cite{pseudolabel, selftraining} integrate unlabeled data via assigning pseudo labels to them, with the knowledge acquired from labeled data. In the era of deep learning, subsequent methods mainly follow such a bootstrapping fashion, but greatly boost it with some key components. Specifically, to enhance the quality of pseudo labels, $\Pi$-model \cite{pimodel} and Mean Teacher \cite{meanteacher} ensemble model predictions and model parameters respectively. Later works start to exploit the role of perturbations. During this trend, \cite{sajjadi2016regularization} proposes to apply stochastic perturbations on inputs or features, and enforce consistency across these predictions. Then, UDA \cite{uda} emphasizes the necessity of strengthening the perturbation pool. It also follows VAT \cite{vat} and MixMatch \cite{mixmatch} to supervise the prediction under strong perturbations with that under weak perturbations. Since then, weak-to-strong consistency regularization has become a standard practice in SSL. Eventually, the milestone work FixMatch \cite{fixmatch} presents a simplified framework using a fixed confidence threshold to discard uncertain samples. Our ShrinkMatch is built upon FixMatch. But, we highlight the value of previously neglected uncertain samples, and leverage them in an informative but also safe manner.

More recently, DST \cite{debiased} decouples the generation and utilization of pseudo labels with a main and an auxiliary head respectively. Besides, SimMatch \cite{simmatch} explores instance-level relationships with labeled embeddings to supplement original class-level relations. Compared with them, our ShrinkMatch achieves larger improvements.

\vspace{2mm}
\noindent
\textbf{Defining uncertain samples.} Earlier works estimate the uncertainty with Bayesian Neural Networks \cite{kendall2017uncertainties}, or its faster approximation, \eg, Monte Carlo Dropout \cite{dropout}. Some other works measure the prediction disagreement among multiple randomly augmented inputs \cite{wang2019aleatoric}. The latest trend is to directly use the entropy of predictions \cite{uamt}, cross entropy \cite{dash}, or softmax confidence \cite{fixmatch} as a measurement for uncertainty. Our work is not aimed at the optimal uncertainty estimation strategy, so we adopt the simplest solution from FixMatch, \ie, using the maximum softmax output as the certainty. 

\vspace{2mm}
\noindent
\textbf{Utilizing uncertain samples.}
UPS \cite{ups} leverages negative class labels whose confidence is below a pre-defined threshold, from a reversed multi-label classification perspective. In comparison, our model is enforced to tell the most likely class without being cheated by the less likely ones. So our supervision on uncertain samples is more informative and produces more discriminative representations. Moreover, we do not introduce any extra hyper-parameters, \eg, the lower threshold in \cite{ups}, into our framework.
\section{Method}

We primarily provide some notations and review a common practice in semi-supervised learning (SSL) (\cref{sec:preliminaries}). Next, we present our ShrinkMatch in detail (\cref{sec:shrink} and \cref{sec:reweight}). Finally, we summarize our approach and provide a further discussion (\cref{sec:summary} and \cref{sec:discussion}).

\subsection{\label{sec:preliminaries}Preliminaries}

Semi-supervised learning aims to learn a model with limited labeled images $\mathcal{D}^l = \{(x_k, y_k)\}$, aided by a large number of unlabeled images $D^u = \{u_k\}$. Recent frameworks commonly follow the FixMatch practice. Concretely, an unlabeled image $u$ is first transformed by a weak augmentation pool $\mathcal{A}^w$ and a strong augmentation pool $\mathcal{A}^s$ to yield a pair of weakly and strongly augmented images $(u^w, u^s)$. Then, they are fed into the model together to produce corresponding predictions $(p^w, p^s)$. Typically, $p^w$ is of much higher accuracy than $p^s$, while $p^s$ is beneficial to learning. Therefore, $p^w$ serves as the pseudo label for $p^s$. Moreover, a core practice introduced by FixMatch is that, to improve the quality of selected pseudo labels, a fixed confidence threshold is pre-defined to abandon uncertain ones. So the unsupervised loss $\mathcal{L}_u$ can be formulated as:
\vspace{-2mm}
\begin{equation}
\vspace{-2mm}
    \mathcal{L}_u = \frac{1}{B_u}\sum_{k=1}^{B_u} \mathbbm{1}(\xi(p^w_k) \geq \tau)\cdot\mathrm{H}(p^w_k, p^s_k),
\label{equ:loss_u}
\end{equation}
where $B^u$ is the batch size of unlabeled images and $\tau$ is the pre-defined threshold. $\xi(p^w_k)$ computes the confidence of logits $p^w_k$ by $\xi(\cdot) = \max(\sigma(\cdot))$, where $\sigma$ is the softmax function. The $\mathrm{H}$ denotes the consistency regularization between the two distributions. It is typically the cross entropy loss.

In addition, the labeled images are learned with a regular cross entropy loss to obtain the supervised loss $\mathcal{L}_x$. The overall loss in each mini-batch then will be:
\vspace{-2mm}
\begin{equation}
\vspace{-2mm}
    \mathcal{L} = \mathcal{L}_x + \lambda_u\cdot\mathcal{L}_u,
\end{equation}
where $\lambda_u$ acts as a trade-off term between the two losses.

\begin{figure*}[t]
    \centering
    \includegraphics[width=0.95\linewidth]{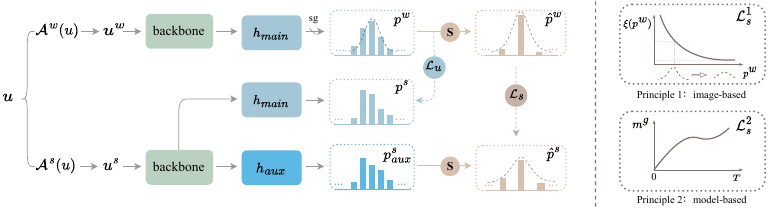}
    \vspace{-2mm}
    \caption{An overview of our proposed ShrinkMatch. Our motivation is to fully leverage the originally uncertain samples. ``\textbf{S}'' denotes \textbf{shrinking the class space}. The confusion classes for the top-1 class are detected and removed \emph{in a fully automatic and instance-adaptive fashion}, to construct a shrunk space where the top-1 class is turned certain. $\mathcal{L}_u$ is the original certain loss, while $\mathcal{L}_s$ calculates the uncertain loss in the shrunk class space. We add an auxiliary head $h_{aux}$ to learn in the new space. On the right, we further reweight $\mathcal{L}_s$ based on two principles. Principle 1 (\textbf{image-based}): image predictions with larger reliability are attached more importance. Principle 2 (\textbf{model-based}): We track the model state during training for reweighting.}
    \vspace{-2mm}
    \label{fig:shrinkmatch}
\end{figure*}

\subsection{\label{sec:shrink}Shrinking the Class Space for Certainty}

\noindent
\textbf{Our motivation.}
As reviewed above, FixMatch discards the samples whose confidence is lower than a pre-defined threshold, because their pseudo labels are empirically found relatively noisier. These samples are named uncertain samples in this work. Although this practice retains high-quality pseudo labels, it incurs a low utilization of the whole unlabeled set, especially when the scenario is challenging and the selection criterion is strict. Take the CIFAR-100 dataset as an instance, with a common threshold of 0.95 and 4 labels per class, there will be nearly 20\% unlabeled samples being ignored due to their low certainty. We argue that, these uncertain samples can still benefit the model optimization, as long as we can design appropriate constraints (loss functions) on them. So we first investigate the cause of low certainty to gain some better intuitions.

The reason for the low certainty of an unlabeled image is that, the model tends to be confused among some top classes. For example, given a cat image, the score of the class \texttt{tabby} \texttt{cat} and class \texttt{tiger} \texttt{cat} may be both high, so the model is not absolutely certain what the concrete class is. Motivated by this observation, we propose to shrink the original class space via adaptively detecting and removing the confusion classes for the top-1 class. Then the shrunk space is only composed of the original top-1 class, as well as the remaining less likely classes. After this process, re-calculated confidence of the top-1 class will satisfy the pre-defined threshold. Thereby, we can enforce the model to learn the previously uncertain samples in this new certain space, as shown in \cref{fig:shrinkmatch}. Following the previous cat example, if the top-1 class is \texttt{tabby} \texttt{cat}, our method will scan scores of all other classes, and remove confusion ones (\eg, \texttt{tiger} \texttt{cat} and \texttt{siamese} \texttt{cat}) to construct a confident shrunk space, where the model is sure that the image is a \texttt{tabby} \texttt{cat}, rather than a \texttt{table}. Since we do not ask the model to discriminate among several top classes, it will avoid suffering from the noise when it makes a wrong judgment in the original space.

\vspace{1mm}
\noindent
\textbf{How to seek the shrunk class space?}
Now, how can we enable our method to \emph{automatically} seek the optimal shrunk space of an uncertain unlabeled image? Ideally, we hope this seeking process is free from any prior knowledge from humans, \eg, class relationships, and also does not require any extra hyper-parameters. Considering these, we opt to inherit the pre-defined confidence threshold as a criterion. To be specific, we first \emph{sort} the predicted logits in a descending order to obtain $p^w = \{s^w_{n_i}\}_{i=1}^C$ for classes $\{n_i\}_{i=1}^C$, where $s^w_{n_i} \ge s^w_{n_{i+1}}$. In the shrunk space, we will retain the original top-1 class, because it is still the most likely one to be true. Then, we find a set of less likely classes $\{n_i\}_{i=K}^C$ by enforcing two constraints on the $K$:
\begin{align}
&\xi(\{s^w_{n_1}\} \cup \{s^w_{n_i}\}_{i=K}^C) \ge \tau\label{equ:certain},\\
&\xi(\{s^w_{n_1}\} \cup \{s^w_{n_i}\}_{i=K-1}^C) < \tau\label{equ:largest},
\end{align}
where $\xi$ is defined the same as that in \cref{equ:loss_u}, calculating the confidence of the re-assembled logits. The final shrunk space is composed of the re-assembled classes $\{n_1\} \cup \{n_i\}_{i=K}^C$. The two constraints on $K$ not only ensure the top-1 class is turned certain in the new space (\cref{equ:certain}), but also select the largest space among all candidates (\cref{equ:largest}).

\vspace{1mm}
\noindent
\textbf{How to learn in the shrunk class space?} For a \emph{weakly} augmented uncertain image $x^w$, the model is certain about its top-1 class in the shrunk class space. To learn effectively in this space, we follow the popular practice of weak-to-strong consistency regularization. The \emph{correspondingly shrunk} prediction on the \emph{strongly} augmented image is enforced to match that on the \emph{weakly} augmented one. Concretely, for clarity, the re-assembled logits from $p^w$ is denoted as $\colorhat{p}^w$, which means $\colorhat{p}^w = \{s^w_{n_1}\} \cup \{s^w_{n_i}\}_K^C$. We use the re-assembled classes $\{n_1\} \cup \{n_i\}_K^C$ in the shrunk space to correspondingly gather the logits $p^s$ on $x^s$, yielding $\colorhat{p}^s = \{s^s_{n_1}\} \cup \{s^s_{n_i}\}_K^C$. Then we can regularize the consistency between the two shrunk distributions $\colorhat{p}^s$ and $\colorhat{p}^w$, similar to that in \cref{equ:loss_u}:
\begin{equation}
    \mathcal{L}_s = \frac{1}{B_u}\sum_{k=1}^{B_u} \mathbbm{1}(\xi(p^w_k) < \tau)\cdot\colorhat{\mathrm{H}}(\colorhat{p}^w_k, \colorhat{p}^s_k),
\end{equation}
where the indicator function is to find uncertain samples.

Empirically, we observe that if we use the original linear head $h_{main}$ to learn this auxiliary supervision, it will make the confidence of our model increase aggressively. Most \emph{noisy} unlabeled samples are blindly judged as certain ones. We conjecture that it is because the $\mathcal{L}_s$ strengthens weights of the classes that are frequently uncertain, then these classes will be incorrectly turned certain. With this in mind, our solution is simple. We adopt an auxiliary MLP head $h_{aux}$ that shares the backbone with $h_{main}$, to deal with this auxiliary optimization target, as shown in \cref{fig:shrinkmatch}. So the $\colorhat{p}^s$ is indeed gathered from predictions of $h_{aux}$ ($\colorhat{p}^w$ is still from $h_{main}$). This modification enables our feature extractor to acquire more discriminative representations, and meantime does not affect predictions of the main head. Note that $h_{aux}$ is only applied for training, bringing no burden to the test stage.

\subsection{\label{sec:reweight}Reweighting the Uncertain Loss}

Despite the effectiveness of the above uncertain loss, there still exist two main drawbacks. (1) On one hand, it overlooks the varied reliability of the top-1 class among different uncertain images. For example, suppose two uncertain images $u_1$ and $u_2$ with softmax predictions [0.8, 0.1, 0.1] and [0.5, 0.3, 0.2] in the original space, they should not be treated equally in the shrunk space. The $u_1$ with top-1 confidence of $0.8$ is more likely to be true than $u_2$, and thereby should be attached more attention to. (2) On the other, it ignores the gradually improved model performance as the training proceeds. To be concrete, at the very start of training, the predictions are extremely noisy or even random. And then at later stages, the predictions become more and more reliable. So the uncertain predictions at different training stages should not be treated equally. Therefore, we further design two reweighting principles for the two concerns.

\vspace{1mm}
\noindent
\textbf{Principle 1: Reweighting with image-based varied reliability.} According to the above intuition, we directly reweight the uncertain loss of each uncertain image by its top-1 confidence $\xi(p^w_k)$ in the \emph{original} class space, which is:
\vspace{-1mm}
\begin{equation}
\vspace{-1mm}
    \mathcal{L}_s^1 = \frac{1}{B_u}\sum_{k=1}^{B_u} \mathbbm{1}(\xi(p^w_k) < \tau)\cdot\colorhat{\mathrm{H}}(\colorhat{p}^w_k, \colorhat{p}^s_k)\cdot\xi(p^w_k).
\label{eq:raw_uncertain}
\end{equation}

We do not use the top-1 confidence $\xi(\colorhat{p}^w_k)$ in the \emph{shrunk} space as the weight, because after shrinking, this value of different predictions is very close to each other. So generally, $\xi(p^w_k)$ is more discriminative than $\xi(\colorhat{p}^w_k)$ as the weight.

\vspace{1mm}
\noindent
\textbf{Principle 2: Reweighting with model-based gradually improved state.} One na\"ive solution is to linearly increase the loss weight of $\mathcal{L}_s$ from 0 at the beginning to $\mu$ at iteration $T$, and then keep $\mu$ until the end. However, this practice has two severe disadvantages. First, the two additional hyper-parameters $\mu$ and $T$ are not easy to determine, and could be sensitive. More importantly, the linear scheduling criterion simply assumes the model state also improves linearly. Indeed, it can not be true. Thus, we here present a more promising principle to perform reweighting, that is free from any extra hyper-parameters. It can adjust the loss weight according to the model state in a fully adaptive fashion. To be specific, we use the certain ratio of the unlabeled set as an indicator for the model state. The certain ratio is traced at each iteration and accumulated globally in an exponential moving average (EMA) manner. Formally, the certain ratio in a single mini-batch is given by:
\begin{equation}
\vspace{-2mm}
    m = \frac{1}{B_u}\sum_{k=1}^{B_u} \mathbbm{1}(\xi(p^w_k) \ge \tau).
\end{equation}

The \emph{global} certain ratio $m^g$ is initialized as 0, and accumulated at each training iteration by:
\begin{equation}
    m^g \leftarrow \gamma\cdot m^g + (1 - \gamma)\cdot m,
\end{equation}
where $\gamma$ is the momentum coefficient. It is a hyper-parameter already defined in FixMatch (baseline), where it is used to update the teacher parameters for final evaluation.

Obviously, $m^g$ falls between $0$ and $1$. And it will approximately increase from $0$ to a nearly saturated value. Then, the reweighted uncertain loss is given by:
\vspace{-2mm}
\begin{equation}
\vspace{-2mm}
    \mathcal{L}_s^2 = m^g\cdot \frac{1}{B_u}\sum_{k=1}^{B_u} \mathbbm{1}(\xi(p^w_k) < \tau)\cdot\colorhat{\mathrm{H}}(\colorhat{p}^w_k, \colorhat{p}^s_k).
\end{equation}

Integrating the above two intuitions and principles, the final reweighted uncertain loss will be:
\begin{equation}
\label{eq:uncertain}
    \mathcal{L}_s = m^g\cdot \frac{1}{B_u}\sum_{k=1}^{B_u} \mathbbm{1}(\xi(p^w_k) < \tau)\cdot\colorhat{\mathrm{H}}(\colorhat{p}^w_k, \colorhat{p}^s_k)\cdot\xi(p^w_k).
\end{equation}

\subsection{\label{sec:summary}Summary}

To summarize, the final loss in a mini-batch is a combination of the supervised loss ($\mathcal{L}_x$), certain loss ($\mathcal{L}_u$, \cref{equ:loss_u}), and uncertain loss in the shrunk class space ($\mathcal{L}_s$, \cref{eq:uncertain}):
\begin{equation}
    \mathcal{L} = \mathcal{L}_x + \lambda_u\cdot(\mathcal{L}_u + \mathcal{L}_s).
\end{equation}

We do not carefully fine-tune the fusion weight between $\mathcal{L}_u$ and $\mathcal{L}_s$ , but use 1:1 by default to avoid hyper-parameters.

\subsection{\label{sec:discussion}Discussions}

Our uncertain loss in the shrunk space owns two properties: \textbf{informative} and also \textbf{safe}. The first property is because we manage to find the largest shrunk space that reaches the confidence threshold. Besides, we also adopt the weak-to-strong consistency regularization to pose a challenging optimization target. Both constraints ensure the learning in the shrunk space is not trivial and still quite informative. Next, we wish to explain the second property ``safe'', especially about \emph{how we avoid noise in the shrunk space}.

Noises in pseudo labels distinguish the semi-supervised paradigm from the fully-supervised one. So designing a safe optimization target for unlabeled data is crucial. Typically, the cross entropy loss will maximize the softmax probability $\exp{(s_t)} / \sum_{i=1}^C\exp(s_i) \rightarrow 1$ for the target class $t$. It inevitably suppresses scores of all other classes except $t$. However, the true class could not be $t$, and may be the \nth{2} or \nth{3} largest class, \etc, which is wrongly restrained. This is frequently observed when the confidence of class $t$ is not high enough. As a milder alternative, the soft labeling still encounters a similar problem. In contrast, our shrunk target directly \emph{excludes} these confusion classes. So only the almost unlikely classes are suppressed. 

\textbf{It is worth noting}, even if the predicted top-1 class is not in line with the human label, it is probably one of the closest semantics, \eg, belonging to the same superclass as the groundtruth label, as shown in \cref{fig:superclass}. Thus, contrasting these relevant classes against other less likely classes with our noise-robust shrunk loss is still beneficial. The model is encouraged to make discriminative predictions closer to top classes compared to tail classes. In addition, our disentangled auxiliary head can effectively leverage such supervision while not affecting the main classification tasks \cite{debiased}.
\section{Experiment}

\begin{table}[t]
\centering
\small
\renewcommand{\arraystretch}{1.1}
\setlength\tabcolsep{1.1mm}
    \centering
    \begin{tabular}{l|ccccc|c}
    \toprule
    
    Seed & 0 & 1 & 2 & 3 & 4 & Mean \\
    
    \midrule
    
    SimMatch \cite{simmatch} & 95.34 & 95.16 & 92.63 & 93.76 & 95.10 & 94.39 \\
    
    \textbf{ShrinkMatch} \textbf{| \textcolor{gray}{{\footnotesize{40}}}} & 95.09 & 94.66 & 95.12 & 94.78 & 94.95 & \textbf{94.92} \\
    
    \midrule
    
    SimMatch \cite{simmatch} & 95.58 & 95.50 & 95.34 & 94.06 & 95.26 & 95.15 \\
    
    \textbf{ShrinkMatch} \textbf{| \textcolor{gray}{{\footnotesize{250}}}} & 95.39 & 95.44 & 95.36 & 94.76 & 95.35 & \textbf{95.26} \\
    
    \bottomrule

    \end{tabular}
    \vspace{-1mm}
    \caption{Comparison with SOTAs on \textbf{CIFAR-10}. The same seed ensures exactly the same data split. The \textcolor{gray}{\textbf{400}} or \textcolor{gray}{\textbf{2500}} denotes the number of labels.}
    \vspace{-2mm}
    \label{tab:cifar10}
\end{table}

In this section, we first describe the implementation details of our framework. Then, we compare our ShrinkMatch with previous state-of-the-art methods (SOTAs) under extensive evaluation protocols. Lastly, we conduct comprehensive ablation studies on each component to validate the necessity.

\subsection{Implementation Details}

\vspace{1mm}
\noindent
\textbf{Baselines.}
We use FixMatch + distribution alignment (DA) as our baseline on all datasets except ImageNet and SVHN. On ImageNet, we build our method on SimMatch. On SVHN, we discard DA. To be more convincing, we adopt the same codebase as our compared methods.

\vspace{1mm}
\noindent
\textbf{Hyper-parameters.} Following prior arts, Wide ResNet-28-2  \cite{wrn}, WRN-28-8, WRN-37-2, and WRN-28-2 are used for CIFAR-10, CIFAR-100, STL-10, and SVHN respectively. A ResNet-50 \cite{resnet} is used for ImageNet. The auxiliary head $h_{aux}$ for uncertain samples is a 3-layer MLP. On the ImageNet, we set $B_u$ = $64\times5$, but for other datasets, $B_u$ = $64\times7$. The labeled batch size is $64$ for all datasets. On the ImageNet, the model is trained for $400$ epochs, while on the others, it is trained for $2^{20}$ iterations. The initial learning rate is set as $0.03$ for all datasets with a cosine scheduler. Specially, on the ImageNet, it is warmed up for $5$ epochs. The consistency regularization $\textrm{H}$ in $\mathcal{L}_u$ on STL-10 and SVHN is a hard cross entropy (CE) loss, while on the CIFAR-10/100 and ImageNet, following the SimMatch and CoMatch \cite{comatch}, it is modified to a soft CE loss. The $\colorhat{\textrm{H}}$ in our proposed $\mathcal{L}_s$ is simply a hard CE loss. The weight $\lambda_u$ for the two unsupervised losses is set as $10$ on the ImageNet and $1$ on others. The confidence threshold $\tau$ is $0.7$ on the ImageNet and $0.95$ for others. The shared momentum coefficient $\gamma$ between our global certain ratio $m^g$ and the teacher parameters is $0.999$. Following the common practice, the teacher model is only maintained for final evaluation.

\begin{table}[t]
\centering
\small
\renewcommand{\arraystretch}{1.1}
\setlength\tabcolsep{1.05mm}
    \centering
    \begin{tabular}{l|ccccc|c}
    \toprule
    
    Seed & 0 & 1 & 2 & 3 & 4 & Mean \\
    
    \midrule
    
    SimMatch \cite{simmatch} & 62.06 & 60.19 & 59.89 & 64.88 & 63.92 & 62.19 \\
    
    \textbf{ShrinkMatch} \textbf{| \textcolor{gray}{\footnotesize{400}}} & 65.00 & 63.47 & 63.77 & 66.42 & 64.52 & \textbf{64.64} \\
    
    \midrule
    
    SimMatch \cite{simmatch} & 74.64 & 75.19 & 74.53 & 75.03 & 75.24 & \textbf{74.93} \\
    
    \textbf{ShrinkMatch} \textbf{| \textcolor{gray}{\footnotesize{2500}}} & 75.00 & 75.11 & 74.54 & 74.78 & 74.72 & 74.83 \\
    
    \bottomrule

    \end{tabular}
    \vspace{-1mm}
    \caption{Comparison with SOTAs on \textbf{CIFAR-100}.}
    \vspace{-2mm}
    \label{tab:cifar100}
\end{table}

\begin{table}[t]
\centering
\small
\setlength\tabcolsep{2.9mm}
    \centering
    \begin{tabular}{l|ccc|c}
    \toprule
    
    Seed & 0 & 1 & 2 & Mean \\
    
    \midrule
    
    FixMatch \cite{fixmatch} & 65.85 & 67.94 & 58.30 & 64.03 \\
    
    FlexMatch \cite{flexmatch} & 76.71 & 68.28 & 67.55 & 70.85 \\
    
    \textbf{ShrinkMatch} \textbf{| \textcolor{gray}{\footnotesize{40}}} & 85.75 & 85.64 & 86.55 & \textbf{85.98} \\
    
    \midrule
    
    FixMatch \cite{fixmatch} & 90.91 & 88.71 & 90.94 & 90.19 \\
    
    FlexMatch \cite{flexmatch} & 91.35 & 92.29 & 91.66 & \textbf{91.77} \\
    
    \textbf{ShrinkMatch} \textbf{| \textcolor{gray}{\footnotesize{250}}} & 91.13 & 92.43 & 91.10 & 91.55 \\
    
    \bottomrule

    \end{tabular}
    \vspace{-1mm}
    \caption{Comparison with SOTAs on \textbf{STL-10}.}
    \vspace{-2mm}
    \label{tab:stl10}
\end{table}

\begin{table}[t]
\centering
\small
\setlength\tabcolsep{2.9mm}
    \centering
    \begin{tabular}{l|ccc|c}
    \toprule
    
    Seed & 0 & 1 & 2 & Mean \\
    
    \midrule
    
    FixMatch \cite{fixmatch} & 94.53 & 96.90 & 97.14 & 96.19 \\
    
    FlexMatch \cite{flexmatch} & 89.19 & 89.93 & 96.32 & 91.81 \\
    
    \textbf{ShrinkMatch} \textbf{| \textcolor{gray}{\footnotesize{40}}} & 97.96 & 97.81 & 96.70 & \textbf{97.49} \\
    
    \midrule
    
    FixMatch \cite{fixmatch} & 98.00 & 97.99 & 97.94 & 97.98 \\
    
    FlexMatch \cite{flexmatch} & 91.76 & 91.83 & 96.65 & 93.41 \\
    
    \textbf{ShrinkMatch} \textbf{| \textcolor{gray}{\footnotesize{250}}} & 98.08 & 98.06 & 97.98 & \textbf{98.04} \\
    
    \bottomrule

    \end{tabular}
    \vspace{-1mm}
    \caption{Comparison with SOTAs on \textbf{SVHN}.}
    \vspace{-2mm}
    \label{tab:svhn}
\end{table}

\begin{table*}[t]
\centering
\small
\setlength\tabcolsep{3.5mm}
    \centering
    \begin{tabular}{llcc|cccc}
    \toprule
    
    \multirow{2}{*}{Pre-training} & \multirow{2}{*}{Method} & \multirow{2}{*}{Epochs} & Params & \multicolumn{2}{c}{\textbf{Top-1}} & \multicolumn{2}{c}{\textbf{Top-5}} \\
    
    ~ & ~ & ~ & (train / test) & 1\% & 10\% & 1\% & 10\% \\
    
    \midrule
    
    SimCLR v2 \cite{simclrv2} & \multirow{3}{*}{Fine-tune} & 800 & 34.2M / 25.6M & 57.9 & 68.4 & 82.5 & 89.2 \\
    
    SwAV \cite{swav} & ~ & 800 & 30.4M / 25.6M & 53.9 & 70.2 & 78.5 & 89.9 \\
    
    WCL \cite{wcl} & ~ & 800 & 34.2M / 25.6M & 65.0 & 72.0 & 86.3 & 91.2 \\
    
    \cmidrule{2-8}
    
    \multirow{2}{*}{MoCo v2 \cite{mocov2}} & Fine-tune & 800 & 30.0M / 25.6M & 49.8 & 66.1 & 77.2 & 87.9 \\
    
    ~ & CoMatch \cite{comatch} & 1200 & 30.0M / 25.6M & 67.1 & 73.7 & 87.1 & 91.4 \\
    
    MoCo-EMAN \cite{eman} & FixMatch-EMAN \cite{eman} & 1100 & 30.0M / 25.6M & 63.0 & 74.0 & 83.4 & 90.9 \\
    
    \midrule
    
    \multirow{3}{*}{None} & CoMatch \cite{comatch} & 400 & 30.0M / 25.6M & 66.0 & 73.6 & 86.4 & 91.6 \\
    
    ~ & SimMatch$^\dag$ \cite{simmatch} & 400 & 30.0M / 25.6M & 67.0 & 74.1 & 86.9 & 91.5 \\
    
    ~ & \textbf{ShrinkMatch} & 400 & 31.8M / 25.6M & \textbf{67.5} & \textbf{74.5} & \textbf{87.4} & \textbf{91.9} \\
    
    \bottomrule

    \end{tabular}
    \vspace{-1mm}
    \caption{Accuracy (\%) on the \textbf{ImageNet-1K} with 1\% and 10\% labeled images. $\dag$: Reproduced in our environment.}
    \label{tab:imagenet}
\end{table*}

\begin{figure*}[t]
\centering
    \begin{minipage}{0.31\linewidth}
    \centering\captionsetup[subfigure]{justification=centering}
    \includegraphics[width=\linewidth]{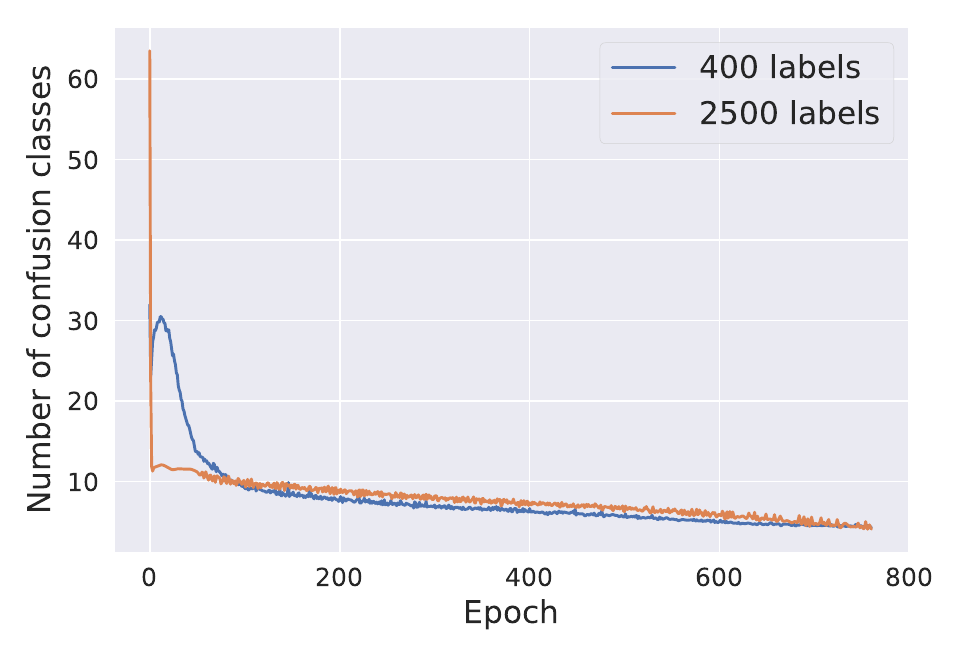}
    \subcaption{Number of removed confusion classes.}
    \label{fig:n_confusion}
    \end{minipage}
    \hspace{2mm}
    \begin{minipage}{0.31\linewidth}
    \includegraphics[width=\linewidth]{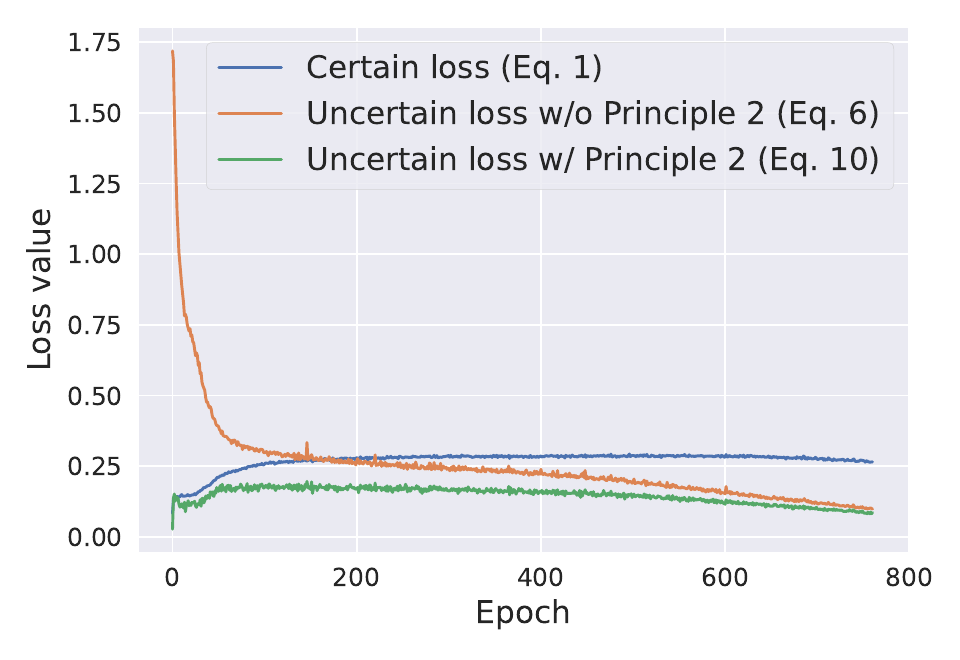}
    \subcaption{Certain loss $\mathcal{L}_u$ and uncertain loss $\mathcal{L}_s$.}
    \label{fig:loss}
    \end{minipage}
    \hspace{2mm}
    \begin{minipage}{0.31\linewidth}
    \includegraphics[width=\linewidth]{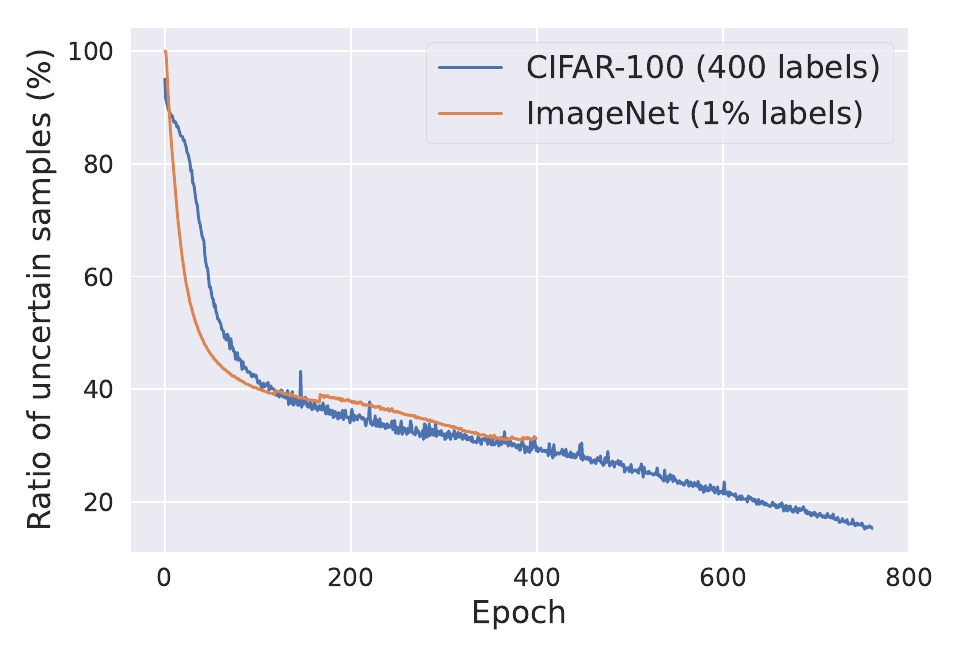}
    \subcaption{Ratio of uncertain samples.}
    \label{fig:proportion}
    \end{minipage}
    \vspace{-2mm}
\caption{(a) The number of removed confusion classes for each uncertain image as the training proceeds. (b) Value of different loss functions, \ie, \cref{equ:loss_u}, \cref{eq:raw_uncertain}, and \cref{eq:uncertain}. (c) The ratio of uncertain samples (certainty $< \tau$) in each mini-batch.}
\vspace{-2mm}
\end{figure*}

\subsection{CIFAR-10 and CIFAR-100}

The CIFAR dataset is composed of 50000/10000 training/validation images of size $32\times32$. The CIFAR-10 contains 10 balanced classes to recognize, while CIFAR-100 is much more challenging, containing 100 classes. We run our ShrinkMatch on five different seeds and report each result as well as the average result. On the CIFAR-10 of \cref{tab:cifar10} where performance is almost saturated, we still obtain a non-trivial improvement of 0.53\% (94.39\% $\rightarrow$ 94.92\%) with 4 labels per class (400 labels in total). In addition, notably, as shown in \cref{tab:cifar100}, with only 4 labels per class on the challenging CIFAR-100, our ShrinkMatch remarkably outperforms SimMatch by 2.45\% (62.19\% $\rightarrow$ 64.64\%) on average.

\subsection{STL-10 and SVHN}

The STL-10 dataset originally consists of 5K/100K/8K labeled/unlabeled/validation images. We follow FlexMatch to only select a subset of 40/250 labeled images, while the unlabeled set remains unchanged. As shown in \cref{tab:stl10}, with 4 labels per class, our ShrinkMatch surpasses FlexMatch tremendously from 70.85\% to 85.98\% (+15.13\%).

The SVHN dataset is suitable to reveal the ability of different semi-supervised algorithms in the presence of class imbalance issue. As demonstrated in \cref{tab:svhn}, although the performance of the original FixMatch (our baseline) has almost touched the upper bound, we still further boost it from 96.19\% to 97.49\% (+1.30\%) with 4 labels per class.

\subsection{ImageNet-1K}

The ImageNet dataset is rather challenging, containing 1.28M/50K training/validation images of 1000 classes. We exactly follow the codebase of SimMatch \cite{simmatch}. As shown in \cref{tab:imagenet}, our ShrinkMatch further boosts previous best results on both settings of 1\% and 10\% labeled images.

\begin{table}[t]
\centering
\small
\setlength\tabcolsep{1.5mm}
    \centering
    \begin{tabular}{l|ccccc|c}
    \toprule
    
    Seed & 0 & 1 & 2 & 3 & 4 & Mean \\
    
    \midrule
    
    Baseline & 64.40 & 57.51 & 63.07 & 64.77 & 62.53 & 62.46 \\
    
    \textbf{ShrinkMatch} & 65.00 & 63.47 & 63.77 & 66.42 & 64.52 & \textbf{64.64} \\
    
    \bottomrule

    \end{tabular}
    \vspace{-1mm}
    \caption{Ablation studies of ShrinkMatch on \textbf{CIFAR-100}.}
    \vspace{-2mm}
    \label{tab:ablation_shrinkmatch_cifar100}
\end{table}

\subsection{Ablation Studies}

Unless otherwise specified, we conduct our ablation studies on CIFAR-100 with 4 labels per class.

\vspace{1mm}
\noindent
\textbf{Effectiveness of our holistic ShrinkMatch.} We first view our ShrinkMatch as a holistic component added to our baseline. As displayed in \cref{tab:ablation_shrinkmatch_cifar100} for CIFAR-100, our ShrinkMatch boosts the strong baseline significantly by 2.18\% (62.46\% $\rightarrow$ 64.64\%). And on the STL-10 of \cref{tab:ablation_shrinkmatch_stl10}, the baseline is also improved evidently from 84.85\% to 85.98\% (+1.13\%).

\begin{figure}
    \centering
    \includegraphics[width=0.95\linewidth]{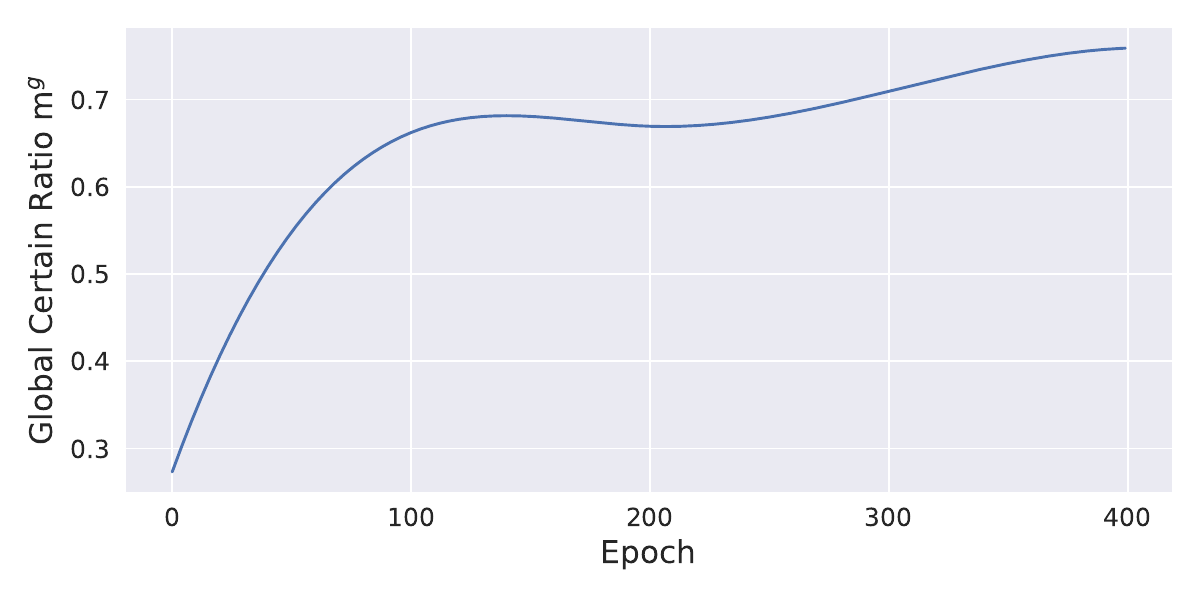}
    \vspace{-2mm}
    \caption{Visualization of our global certain ratio $m^g$.}
    \vspace{-2mm}
    \label{fig:mg}
\end{figure}

\vspace{1mm}
\noindent
\textbf{Effectiveness of two reweighting principles.} In \cref{sec:reweight}, considering the varied prediction reliability of different uncertain samples and different training stages, we propose two principles to reweight our uncertain loss $\mathcal{L}_s$. So here we carefully examine their necessity in \cref{tab:ablation_reweight}. It can be observed that the two principles are both beneficial, jointly improving the basic shrinking practice from 63.16\% to 64.24\%. We also attempt a simple alternative of linearly increasing the uncertain loss weight from 0 (start) to 1 (end). But as evidenced in the Exp 3 \& 4 of \cref{tab:ablation_reweight}, it is obviously inferior to our designed reweighting strategy. We visualize our $m^g$ as training proceeds on ImageNet in \cref{fig:mg}. It looks essentially different from linear scheduling.  Moreover, we visualize the uncertain loss value with or without the second reweighting principle in \cref{fig:loss}. After reweighting, the uncertain loss is of a similar magnitude as the certain loss, which can avoid dominating the gradient at early training stages.

\begin{table}[t]
\small
\centering
\setlength\tabcolsep{3mm}
    \begin{tabular}{l|ccc|c}
    \toprule
    
    Seed & 0 & 1 & 2 & Mean \\
    
    \midrule
    
    Baseline & 84.86 & 84.21 & 85.47 & 84.85 \\
    
    \textbf{ShrinkMatch} & 85.75 & 85.64 & 86.55 & \textbf{85.98} \\
    
    \bottomrule

    \end{tabular}
    \vspace{-1mm}
    \caption{Ablation studies of ShrinkMatch on \textbf{STL-10}.}
    \vspace{-2mm}
    \label{tab:ablation_shrinkmatch_stl10}
\end{table}

\begin{table}[t]
\centering
\small
\setlength\tabcolsep{1.5mm}
    \centering
    \begin{tabular}{c|ccc|cc|c}
    \toprule
    
    Exp & \textbf{Principle 1} & \textbf{Principle 2} & \textbf{LS} & Seed 0 & Seed 1 & Mean \\
    
    \midrule
    
    1 & & & & 65.62 & 60.70 & 63.16 \\
    
    2 & \cmark & & & 65.68 & 62.08 & 63.88 \\
    
    3 & & \cmark & & 65.09 & 62.90 & 64.00 \\
    
    4 & & & \cmark & 63.01 & 59.45 & 61.23 \\
    
    5 & \cmark & \cmark & & 65.00 & 63.47 & \textbf{64.24} \\
    
    \bottomrule

    \end{tabular}
    \vspace{-1mm}
    \caption{Ablation studies on the effectiveness of the two reweighting principles. Due to randomness, the ``Mean'' results are more convincing. \textbf{LS} is short for linear scheduling, which linearly increases the loss weight.}
    \vspace{-2mm}
    \label{tab:ablation_reweight}
\end{table}

\begin{table}[t]
\centering
\small
\setlength\tabcolsep{1.6mm}
    \centering
    \begin{tabular}{c|ccc|cc|c}
    \toprule
    
    Exp & \textbf{Shrink} & \textbf{Aux Head} & \textbf{Label} & Seed 0 & Seed 1 & Mean \\
    
    \midrule
    
    1 & \multicolumn{3}{c|}{Our Strong Baseline} & 64.40 & 57.51 & 60.96 \\
    
    \midrule
    
    2 & \cmark & \cmark & H & 65.00 & 63.47 & \textbf{64.24} \\
    
    3 & \cmark & \cmark & S & 62.92 & 59.76 & 61.34 \\
    
    \midrule
    
    4 & \cmark &  & H & 62.43 & 58.86 & 60.65 \\

    5 & \cmark &  & S & 60.92 & 60.22 & 60.57 \\

    \midrule
    
    6 & & \cmark & H & 64.21 & 59.17 & 61.69 \\
    
    7 & & \cmark & S & 64.75 & 60.89 & 62.82 \\

    \midrule
    
    8 & & & H & 52.90 & 51.88 & 52.39 \\

    9 & & & S & 54.10 & 55.75 & 54.93 \\
    
    \bottomrule

    \end{tabular}
    \vspace{-1mm}
    \caption{Ablation studies on different options to learn uncertain sample. (1) whether to shrink the class space (\textbf{Shrink}), (2) whether to use an auxiliary head (\textbf{Aux Head}), and (3) use hard (\textbf{H}) or soft label (\textbf{S}). Exp 1 is our baseline without using uncertain images. Exp 8 \& 9 are experiments of directly learning uncertain samples, which are quite noisy.}
    \vspace{-2mm}
    \label{tab:ablation_aux}
\end{table}

\vspace{1mm}
\noindent
\textbf{The number of removed confusion classes.}
As introduced before, our approach detects and removes the confusion classes for the top-1 class in a fully automatic and instance-adaptive fashion. So we visualize the number of removed confusion classes for each uncertain sample in \cref{fig:n_confusion}. At the very start of training, model predictions are almost uniform, so abundant classes are removed for an uncertain sample. But as the training proceeds, much fewer confusion classes need to be removed to form a confident shrunk class space.

\vspace{1mm}
\noindent
\textbf{Hard label or soft label in the shrunk class space?} By default, we use the hard label in the shrunk space, following FixMatch. But we also ablate the choice of using the soft label as supervision. As shown in Exp 2 \& 3 of \cref{tab:ablation_aux}, the hard label is much better than the soft label. We conjecture that this is because we have shrunk the class space to a safe one. The soft label will incur some unnecessary noise.

\vspace{1mm}
\noindent
\textbf{Effects of the auxiliary head.}
To prevent the main classifier from being affected, we use an auxiliary head to learn the uncertain samples in the shrunk class space only for discriminative representations. So we first provide the results when still using the main head for the shrunk space. As shown in Exp 2 \& 4 of \cref{tab:ablation_aux}, the auxiliary head is indispensable for our shrinking practice. So we continue to figure out whether the improvement of our ShrinkMatch merely comes from the auxiliary head. We attempt to use the auxiliary head to directly learn the uncertain samples in the original space without shrinking. It can be concluded from Exp 2 \& 6 \& 7 of \cref{tab:ablation_aux}, the original space is indeed much inferior to our shrunk space for uncertain samples, although the auxiliary head still helps the baseline to some extent. Lastly, in \cref{tab:ablation_aux}, we also include two additional results for readers, \ie, directly learning uncertain samples without shrinking or auxiliary head (Exp 8 \& 9), which are quite terrible because of introducing the abundant noise to our main classifier.

\begin{table}[t]
\centering
\small
\setlength\tabcolsep{2.4mm}
    \centering
    \begin{tabular}{l|ccccc}
    \toprule
    
    Threshold & 0.98 & 0.95 & 0.90 & 0.80 & 0.70 \\
    
    \midrule
    
    ShrinkMatch & \textbf{65.80} & 65.00 & 64.55 & 62.74 & 59.58 \\
    
    \bottomrule

    \end{tabular}
    \vspace{-1mm}
    \caption{Ablation studies of the confidence threshold.}
    \vspace{-2mm}
    \label{tab:ablation_threshold}
\end{table}

\vspace{1mm}
\noindent
\textbf{The ratio of uncertain samples.} To further justify our motivation, in \cref{fig:proportion}, we display the ratio of uncertain samples in each mini-batch. It can be seen that even in the middle of the whole training course, there are still around 40\% and 30\% uncertain samples on ImageNet and CIFAR-100 respectively. Therefore, an appropriate approach to utilizing these samples is necessary and definitely beneficial.

\vspace{1mm}
\noindent
\textbf{Different confidence thresholds $\tau$.} We also try different thresholds for uncertain samples, as shown in \cref{tab:ablation_threshold}. When increasing $\tau$ from widely adopted 0.95 to 0.98 (more uncertain samples), our ShrinkMatch can be further improved. These results clearly highlight the effective and safe utilization of uncertain samples with our ShrinkMatch.
\vspace{2.5mm}
\section{Conclusion}

In this work, we aim to fully leverage the uncertain samples in semi-supervised learning. We point out that the low certainty is typically caused by a small portion of confusion classes. Invoked by this, we propose a novel method dubbed ShrinkMatch to automatically detect and remove the confusion classes to construct a shrunk class space, where the top-1 class is turned certain. A weak-to-strong consistency regularization is enforced in the confident new space. Furthermore, we design two reweighting principles for the auxiliary uncertain loss, according to the reliability of different uncertain samples and the gradually improved state of the model. Consequently, our method establishes new state-of-the-art results on widely acknowledged benchmarks.

\paragraph{Acknowledgements.} This work is supported in part by the National Natural Science Foundation of China (62201484, 62222604, 62192783, 62206052), HKU Startup Fund, and HKU Seed Fund for Basic Research.

\clearpage

{\small
\bibliographystyle{ieee_fullname}
\bibliography{egbib}
}

\end{document}